\documentclass[runningheads]{llncs}
\usepackage[T1]{fontenc}
\usepackage{graphicx}
\usepackage{amsmath}
\usepackage{amssymb}
\usepackage{booktabs}
\usepackage{tabularx}
\usepackage{hyperref}
\usepackage{microtype}

\begin{document}

\title{Breaking the Central Bias: Spatially Partitioned Experts for Coordinate-Based Neuroevolution\thanks{R. Claret and A. Gygax contributed equally. R. Claret is the corresponding author. Accepted for publication in the proceedings of the BIOMAP workshop at ICPR 2026 (Springer LNCS). This is the authors' accepted version. The final authenticated publication will be available online via Springer once its DOI is assigned.}}
\titlerunning{Breaking the Central Bias}
\author{Romain Claret\inst{1}\orcidID{0000-0002-5612-8471} \and
Arthur Gygax\inst{1}\orcidID{0009-0002-7154-5537} \and
Michael O'Neill\inst{2}\orcidID{0000-0001-8734-417X} \and
Paul Cotofrei\inst{1}\orcidID{0000-0002-4103-5467} \and
Michael Palma Mendes\inst{1}\orcidID{0009-0008-9964-1057} \and
\\Pascal Felber\inst{1}\orcidID{0000-0003-1574-6721}}
\authorrunning{R. Claret et al.}
\institute{University of Neuch\^{a}tel, Neuch\^{a}tel, Switzerland\\
\email{\{romain.claret, arthur.gygax, paul.cotofrei, michael.palma, pascal.felber\}@unine.ch} \and
University College Dublin, Ireland\\
\email{m.oneill@ucd.ie}}

\maketitle

\begin{abstract}
Evolvable-Substrate HyperNEAT (ES-HyperNEAT), a bio-inspired indirect encoding that determines neuron placement and connection weights from spatial coordinates, exhibits a failure mode on MNIST as a diagnostic benchmark. Because input pixels map to a coordinate space centered at the origin, evolved networks converge on a small central cluster of input pixels, a \textit{spatial-concentration bias}; prior work observed only 21\% mean accuracy in this regime. Is this bias an optimization artifact or an architectural ceiling? Inspired by Mixture-of-Experts (MoE) principles, we partition the input into non-overlapping spatial segments, each assigned to a separately evolved specialist network. With 13 such experts, this design reaches 43\% mean accuracy, a 106\% relative improvement over the baseline. The architectural gain does not depend on data-driven aggregation: equal-weighted averaging, which uses no validation data, already yields a 70\% improvement; the gain comes from partitioning, not the weighting. Receptive-field analysis shows the mechanism: partitioning forces evolution to discover features across the entire image, expanding active pixel coverage from 4\% to 79\%. Absolute accuracy stays below gradient-trained baselines, but the relative gain points to central bias, not the evolutionary search. Two tools are designed to generalize beyond MNIST: a receptive-field diagnostic for silent input-coverage collapse, and a spatial-partitioning remedy that restores coverage.

\end{abstract}

\section{Introduction}

Bio-inspired methods for pattern recognition, such as evolutionary computation and indirect encodings, promise architectures that emerge from a generative process rather than being hand-designed. One failure mode these emergent architectures can exhibit is spatial-concentration bias: the network concentrates on a small, often central, image region and never uses features located elsewhere. Coordinate-based indirect encodings, anchored at the substrate center, can be particularly susceptible.

We study a controlled instance of this failure mode in coordinate-based neuroevolution. Evolvable-Substrate HyperNEAT (ES-HyperNEAT)~\cite{risi2012enhanced} evolves a compressed, geometric representation of connectivity rather than optimizing weights directly (Section~\ref{sec:background}). We use MNIST not as a competitive benchmark but as a diagnostic tool: prior work~\cite{claret2024investigating} showed that ES-HyperNEAT achieved only 20.95\% mean accuracy over 30 runs on MNIST digit classification, and the highest-performing individuals used only a small, centrally located subset of the 784 available pixels, a clean instance of the spatial-concentration failure mode introduced above. This ceiling does not stem from insufficient search: data-driven early stopping makes the search more efficient without raising it~\cite{claret2026pruner}. Two explanations are possible: evolution found an optimal strategy of ignoring peripheral pixels, or the monolithic architecture cannot integrate information across a wide receptive field.

We test the second explanation with a divide-and-conquer intervention. Inspired by Mixture-of-Experts (MoE), we partition the input image into non-overlapping segments, each assigned to a dedicated ``expert.'' Unlike classical MoE with learned gating~\cite{jacobs1991adaptive}, our approach uses deterministic spatial routing, making it a spatially partitioned ensemble. We compare two implementations: an \textit{Independent MoE}, where separate experts evolve as specialists on their respective input slices, and a \textit{Shared MoE}, where a single network processes each segment sequentially. Combining their predictions then requires an aggregation strategy; we evaluate several.

We find a sharp dissociation: the \textit{Independent MoE} reaches 43.07\% mean accuracy, a 106\% relative increase over the baseline, while the \textit{Shared MoE}, which partitions the input identically but processes each segment through a single non-specialized network, reaches only 26.03\% (a very large effect). The gain comes not from partitioning the image alone (both architectures do) but from evolving an independent specialist for each partition. The limiting factor is therefore the monolithic architecture itself, not the search process. Three experimental controls isolate architectural effects from data-presentation variance (Section~\ref{sec:experimental_setup}).

\section{Background}
\label{sec:background}

\subsection{Neuroevolution, ES-HyperNEAT, and TPE}

ES-HyperNEAT~\cite{risi2012enhanced} is an indirect encoding derived from NEAT~\cite{stanley2002evolving}.
Rather than encoding each connection directly, it evolves a Compositional Pattern-Producing Network (CPPN) that, when queried with the geometric coordinates of two neurons on a substrate, returns the connection weight between them. ES-HyperNEAT determines substrate topology through iterative quadtree decomposition: starting from a uniform coarse grid, the algorithm subdivides regions where CPPN output exhibits high variance across child queries, placing nodes only where the connectivity pattern is non-uniform. A variance threshold governs which regions merit further resolution. Because this initial coarse grid is uniform, any center-clustering of nodes emerges from subsequent variance-driven subdivisions rather than from the initialization. Because coordinates are normalized to a bounded space centered at the origin, this geometric encoding can nonetheless produce a central bias whose presence we observe empirically (Section~\ref{sec:discussion}: all 30 monolithic runs concentrate active pixels within a narrow central band) but whose mechanism is not fully isolated.

The bias most likely arises from three properties that compound, not from any one in isolation. (i) By the standard substrate convention, the coordinate frame is anchored at the image center: input positions are normalized to the range $[-1, 1]$ around an origin that coincides with the geometric middle of the image. (ii) Symmetric CPPN activation primitives such as Gaussian and sine produce structured spatial variance that the quadtree's variance-driven subdivision responds to in ways that depend on the primitive's symmetry properties. (iii) Once a workable connectivity pattern is in place within the central region, and central pixels already suffice (as on MNIST), no fitness signal systematically pushes CPPN bias inputs to shift the response off-center, so the search settles rather than continues.

This bias cannot be cleanly isolated on MNIST, whose centered digits align discriminative pixels with the substrate origin, so the coordinate-system bias and the information-density bias point the same way (Section~\ref{sec:discussion}); partitioning bypasses the question by design, because each expert's input window precludes solutions drawn from outside its assigned region. The limitation is specific to coordinate-based indirect encodings: direct encodings like NEAT assign no geometric coordinates to neurons and so cannot exhibit coordinate-dependent spatial bias, though they face prohibitive scaling on high-dimensional inputs, where the directly encoded genome grows with network size.

Like many machine learning methods, neuroevolutionary systems are sensitive to hyperparameter choice. The baseline study~\cite{claret2024investigating} used the Tree-structured Parzen Estimator (TPE)~\cite{bergstra2011algorithms}, a Bayesian optimization method that models distributions of hyperparameters yielding low vs.\ high objective values, making it more sample-efficient than random or grid search; we reuse its configuration (Section~\ref{sec:experimental_setup}).

\subsection{Mixture-of-Experts and Modularity in Neuroevolution}

The Mixture-of-Experts (MoE) is an established ensemble learning framework, originally introduced by Jacobs et al., that decomposes a complex problem by assigning different regions of the input space to specialized expert models~\cite{jacobs1991adaptive,jordan1994hierarchical}. In its classic form, a trainable gating network routes inputs to the most appropriate expert, and their outputs are combined to form a final prediction.

We partition the MNIST input image spatially, inspired by MoE but with deterministic routing rather than a learned gate. The expert-to-input assignment is fixed, making our approach a spatially partitioned ensemble rather than a classical MoE. The open question is then how to aggregate predictions from these specialized, parallel modules.

\section{Related Work}
\label{sec:related_work}

Modular architectures have repeatedly improved neuroevolution on complex tasks. We apply a spatially partitioned ensemble within ES-HyperNEAT, combining ideas from MoE and cooperative coevolution. It is the evolutionary, indirectly encoded analogue of region- or tile-level specialization used elsewhere in pattern recognition.

The HyperNEAT family scales neuroevolution to large networks by evolving the connection weights of a fixed substrate as a function of node geometry~\cite{stanley2009hypercube}, and ES-HyperNEAT~\cite{risi2012enhanced} extends this to evolve the substrate topology itself. Modularity has been a recurring route to better behavior in the family: Verbancsics and Stanley constrain HyperNEAT connectivity to encourage modular structure~\cite{verbancsics2011constraining}, applied there to internal connectivity rather than, as here, to the input itself. Our spatial partitioning is the input-side analogue of such structural priors. The Mixture-of-Experts idea has re-emerged at scale as the sparsely-gated MoE layer~\cite{shazeer2017outrageously}, where learned gating routes inputs to specialists; we instead fix routing geometrically. In coordinate-based neuroevolution, the ensemble diversity such schemes rely on must be \emph{forced} by partitioning, because unconstrained experts collapse onto the same central pixels.

Applying MoE principles directly to adaptive-substrate neuroevolution (ES-HyperNEAT) is, to our knowledge, new. The approach builds on modularity research: Reisinger et al.~\cite{reisinger2004evolving} showed that evolving reusable modules with Modular NEAT improves search efficiency, and Schrum and Miikkulainen~\cite{schrum2016solving} extended this with MM-NEAT, evolving multi-modal behavior via separate output modules. Those works focused on behavioral specialization; ours applies modularity to the input itself, assigning experts to fixed regions.

The \textit{Independent MoE} also resembles Cooperative Coevolutionary Algorithms (CCEAs)~\cite{potter2000cooperative}, where a problem is decomposed and partial solutions evolve in separate populations before being combined for evaluation. Here, the decomposition is spatial: each expert network is a coevolving sub-population responsible for one segment of the input vector. CCEAs typically face a credit assignment problem; our framework sidesteps it through post-evolution aggregation strategies that weight each expert by its measured performance (Section~\ref{ssec:aggregation}).

\section{Experimental Setup}
\label{sec:experimental_setup}

We evaluate how an architecture inspired by MoE affects ES-HyperNEAT performance and learning dynamics.

Experiments used the PUREPLES framework~\cite{westh2017pureples}, the Python reference implementation of ES-HyperNEAT, which extends NEAT-Python~\cite{McIntyre_neat-python}. All evolutionary trials were run for a fixed duration of 20 generations to match the generations used in prior work~\cite{claret2024investigating} for fair comparison.

\subsection{Task, Baseline, and Core Hypothesis}
The task is MNIST 10-class handwritten-digit classification. The baseline is the monolithic model from prior work~\cite{claret2024investigating}, which processes the full 784-pixel image. That model achieved a maximum accuracy of 29\% and a mean accuracy of 20.95\% over 30 runs. That study found that the best-performing networks used only a small, centrally located subset of pixels.

Under ordinary selection pressure, the network converges to the simplest sufficient solution: ignoring peripheral data. Our hypothesis: this behavior limits performance, and explicit spatial decomposition forces the evolutionary search to discover features across the entire visual field.

\subsection{Architectural Design and Methodological Controls}
We test this hypothesis with two architectures and three experimental controls. In the \textit{Independent MoE}, each partition is processed by a separately evolved network. In the \textit{Shared MoE}, a single evolved network processes all partitions sequentially without receiving partition-identity input.

\subsubsection{Input Partitioning.}
The 28$\times$28 MNIST image is flattened into a 784-pixel vector and partitioned into $N_e$ contiguous, non-overlapping segments, each fed to a corresponding expert.

\begin{figure}[htbp]
    \centering
    \includegraphics[width=0.9\textwidth]{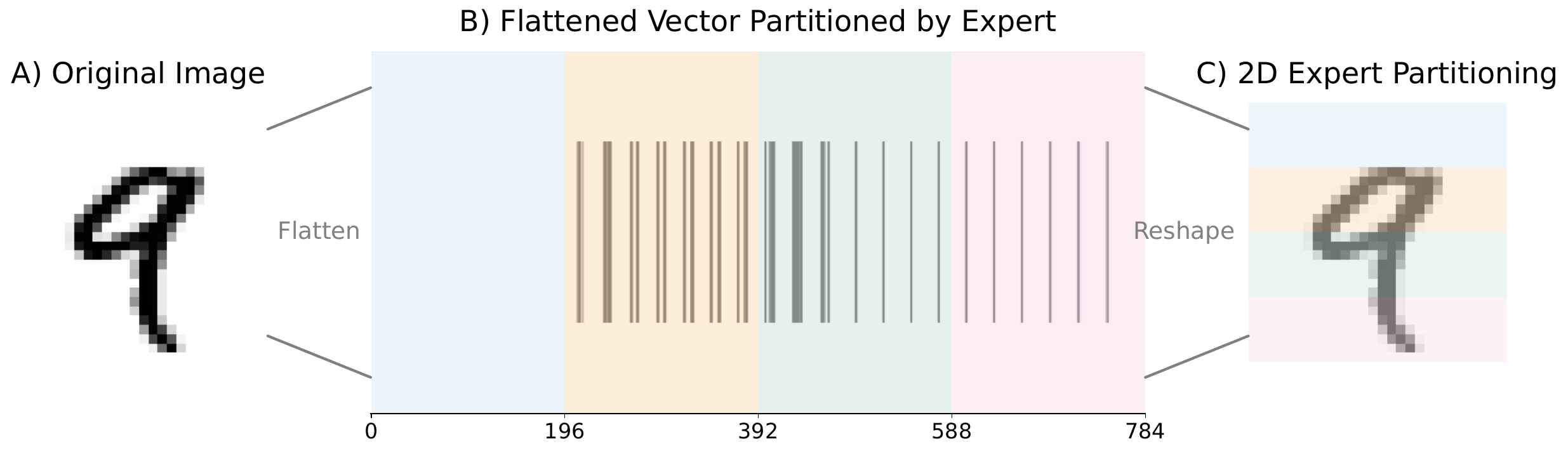}
    \caption{
        Illustration of input partitioning for $N_e=4$ experts.
        \textbf{(A)} The 28$\times$28 image is flattened to 784 pixels and partitioned into four equal, non-overlapping segments.
        \textbf{(B)} Pixel intensities shown as a barcode.
        \textbf{(C)} Reshaping to 2D shows assignment to horizontal slices: Expert 1 (pixels 0--195), Expert 2 (pixels 196--391), Expert 3 (pixels 392--587), Expert 4 (pixels 588--783).
    }
    \label{fig:partitioning}
\end{figure}

To investigate the impact of input granularity, our experiments (both Independent and Shared) explored expert counts from $N_e=1$ to $17$. The bounds of this range are deliberate: $N_e=1$ serves as a direct replication of the monolithic baseline architecture, while $N_e=17$ results in a segment size of approximately 46 pixels. This latter value was chosen to be on the same order of magnitude as the sparsely activated receptive fields observed in the baseline work~\cite{claret2024investigating}, allowing us to test whether forced partitioning at this scale is beneficial. We use a linear partitioning of the input vector, visualized in Figure~\ref{fig:partitioning}.

\subsection{Aggregation Strategies}
\label{ssec:aggregation}
Each expert $e$ produces a score vector $S_e \in \mathbb{R}^{B \times C}$ (batch size $B$, classes $C$). How these vectors are combined into a final prediction determines ensemble quality. We evaluated 14 aggregation strategies across three families; the principal methods of each are described below.

\textbf{Simple Heuristics} (\texttt{Avg}, \texttt{Sum}, \texttt{Max}) average, sum, or take the maximum across expert scores per class, treating all experts equally regardless of accuracy.

\textbf{Mask-Based Heuristics} apply binary or weighted masks $M_e$ based on structural connectivity, silencing experts that cannot vote for certain classes. Three variants are defined: Connection-Masked (\mbox{\texttt{Conn-Masked}}, binary masks), Connection-Weighted (\mbox{\texttt{Conn-Weighted}}, boost-amplified masks), and Threshold-Masked (\mbox{\texttt{Thresh-Masked}}, incorporating connection strength). The aggregated score is:
\begin{equation}
    S_{agg} = \sum_{e=1}^{N_e} (S_e \odot M_e)
\end{equation}
where $\odot$ denotes element-wise multiplication.

\textbf{Performance-Weighted Methods} compute weights $W_{e,c}$ for expert $e$'s competence at predicting class~$c$, derived from per-class $F_1$-scores on a validation set. The \texttt{Perf-\allowbreak{}Weighted} method computes:
\begin{equation}
    S_{agg}[c] = \sum_{e=1}^{N_e} W_{e,c} \cdot S_e[c]
\end{equation}
where $W_{e,c}$ is expert $e$'s $F_1$-score for class~$c$.
Three variants exist:
Structure-Gated (\texttt{Struct-Gated}) requires structural connectivity,
\texttt{Top-Expert} selects the highest-confidence expert,
and Evolved-Weight (\texttt{Evolved-Wt}) augments the per-class $F_1$ weighting with a per-expert importance coefficient (hand-set in the present experiments; intended to be evolved in future work).

The 43.07\% mean and 49\% maximum accuracy were achieved using \texttt{Perf-\allowbreak{}Weighted}. Data-driven aggregation outperforms naive heuristics, though even naive averaging clears the monolithic baseline by a wide margin (Section~\ref{sec:results}).

\subsection{Experimental Control Factors}
Parallelized evolution and input partitioning introduce potential variance. Three control switches isolate the architectural effects:

\begin{itemize}
    \item \textbf{Same-Batch-per-Generation (SBG):} All individuals in a generation are evaluated on an identical, once-drawn batch of images, stabilizing the fitness landscape and removing inter-individual variance from random sampling.
    \item \textbf{Same-Batch-per-Expert (SBE):} Applicable only to the \textit{Independent MoE}, SBE extends SBG so every expert sees the same batch, enabling fair comparison of expert specialization and performance. Enabling SBE automatically enforces SBG.
    \item \textbf{Shared-Mix (SM):} With the \textit{Shared MoE}, this regularizer decorrelates inputs to the shared network by independently shuffling batch indices for each expert slot, breaking the fixed segment-to-slot association and preventing unwanted coadaptation.
\end{itemize}

Default denotes a condition with no control switch applied. These factors yield seven conditions (\textit{Independent MoE} with Default, SBG, or SBE; \textit{Shared MoE} with Default, SBG, SM, or SBG+SM) plus the Monolithic baseline, each run for 30 independent replicates.

\subsubsection{Evolutionary Hyperparameters.}
To isolate the impact of our architectural modifications, all ES-HyperNEAT hyperparameters were fixed across all experimental conditions using the optimal configuration identified in the baseline study~\cite{claret2024investigating}. NEAT: population 100, no initial hidden nodes, connection add/delete probability 0.5, node add/delete probability 0.8/0.2, full direct initial connections (80\% connection fraction), tanh default activation. ES-HyperNEAT substrate: initial depth 2, max depth 5, variance threshold 0.01, division threshold 0.5, max weight 3.0, iteration level 0, tanh activation. Other parameters use the defaults from NEAT-Python's XOR-experiment configuration.

These hyperparameters were optimized for the baseline monolithic architecture, which has access to all 784 input pixels simultaneously. We deliberately reuse them for our models so that any performance gains are attributable to the structural change, not to hyperparameter tuning.

\subsection{Experimental Process and In-depth Analysis}
We first swept all architectural variants (\textit{Independent MoE}, \textit{Shared MoE}) and control factors across the full range of expert counts ($N_e=1$ to $17$). Performance peaked at $N_e=13$ experts with \texttt{Perf-\allowbreak{}Weighted} ($F_1$-based) aggregation. Section~\ref{sec:results} analyzes this configuration and compares all experimental conditions.

\subsection{Analysis Methodology: Tracking Receptive Field Evolution}
\label{ssec:analysis_methodology}
To test whether our architecture changes how networks process the input space, we track the structural evolution of the champion (best-performing individual) at the end of each generation.

From each champion genome, we extract the active receptive field, defined as the set of input pixels that have a connection with a non-zero weight to at least one hidden or output neuron in the phenotype network.

We use this metric to compare how the baseline and our architectures attend to the input space, both quantitatively (receptive-field size over generations) and qualitatively (spatial distribution of active connections).

\section{Experimental Results}
\label{sec:results}

The \textit{Independent MoE} substantially outperforms the monolithic baseline; the \textit{Shared MoE} improves only modestly. We present peak performance, then analyze expert granularity ($N_e$), aggregation strategy, and the structural basis of the difference. Throughout, we report ANOVA $F$-statistics as $F$(numerator df, denominator df) = value and assess pairwise differences with two-sided Welch's t-tests and Cohen's $d$ (0.2/0.5/0.8 = small/medium/large); unless noted otherwise, reported comparisons are significant at $p \le .05$. Our replicated baseline ($N_e=1$) matches the original monolithic model ($t = -0.68$, $p = .502$, $d = -0.17$), confirming a faithful reproduction.

\subsection{Spatially Partitioned Experts Outperform the Monolithic Baseline}
The Default \textit{Independent MoE} with 13 experts achieved 43.07\% mean accuracy, a 106\% relative increase over the 20.95\% monolithic baseline. A two-way ANOVA confirmed that architecture choice significantly affects performance $(F(1, 3536) = 11915.88)$, with the \textit{Independent MoE} outperforming both the \textit{Shared MoE} and the baseline across all conditions. Comparing the best \textit{Independent MoE} against the best \textit{Shared MoE} directly: $t = 29.19$, $d = 7.54$.

The \textit{Shared MoE}'s best variant reached 26.03\% mean accuracy, a 24\% increase over the monolithic model but far below the \textit{Independent} variant. We attribute this gap to a representational bottleneck: sequential processing of disjoint segments without partition identity. Partitioning the input alone is insufficient; how those partitions are processed matters.

\subsection{Impact of Expert Granularity}
\begin{figure}[!htb]
    \centering
    \includegraphics[width=\textwidth]{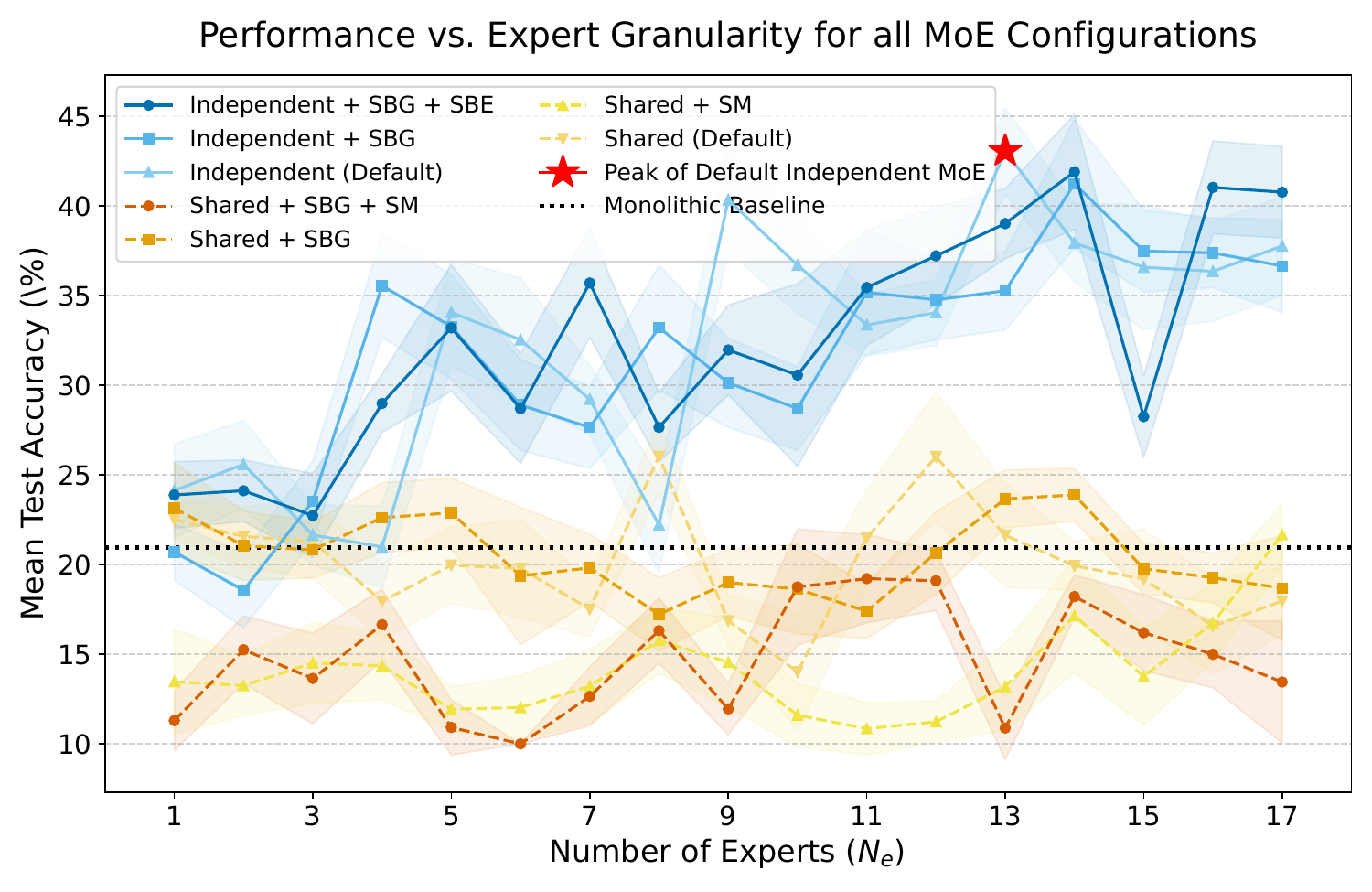}
    \caption{Mean test accuracy as a function of the number of experts ($N_e$). The \textit{Independent MoE} (blue) peaks at $N_e=13$ with 43.07\% mean accuracy; the \textit{Shared MoE} (orange/yellow) shows minimal improvement over the baseline (black dotted), indicating that its architecture does not benefit from increased granularity.}
    \label{fig:accuracy_vs_experts}
\end{figure}

Figure~\ref{fig:accuracy_vs_experts} shows the distinct behaviors of the two architectures across expert counts $N_e=1$ to $17$.

The \textit{Independent MoE} models all trend upward, outperforming the baseline for $N_e > 5$. The interaction between architecture and granularity is highly significant $(F(16, 3536) = 114.43)$: the \textit{Independent MoE}'s performance scales with partitioning, the \textit{Shared MoE}'s does not.

Performance for the top-performing \textit{Independent MoE} (Default) model shows a sharp peak at \textbf{$N_e=13$}. This peak is more than visual: a one-way ANOVA confirmed a significant effect of the number of experts $(F(16, 493) = 230.78)$, and a subsequent Tukey HSD post hoc test revealed that the $N_e=13$ configuration (mean $= 43.07\%$, SD $= 2.45\%$) was statistically superior to all other granularities tested. To measure the magnitude of this improvement, a t-test comparing our best model ($N_e=13$) against our replicated monolithic model ($N_e=1$) confirms a significant difference with a large effect size ($t = 33.69$, $d = 8.70$). This quantifies the benefit of partitioning.

The \textit{Shared MoE} models show no comparable improvement, remaining mostly at or below the 20.95\% baseline. Table~\ref{tab:perf_summary} gives the full statistical breakdown.

The peak position is not arbitrary. At $N_e=13$, each expert receives $784 / 13 \approx 60$ input pixels, on the same order as the $\approx 48$-pixel converged receptive field that the original monolithic baseline~\cite{claret2024investigating} reaches under identical hyperparameter settings. The implication is a practical search-capacity ceiling: ES-HyperNEAT's evolutionary loop, under the 20-generation budget used here, can fully exploit roughly $50$--$60$ input dimensions. Lower expert counts (fewer experts, larger segments) push each expert's segment back above that ceiling, where the central bias likely resurfaces; higher expert counts (more experts, smaller segments) leave too few pixels per expert to support discriminative features.

The \textit{Independent MoE}'s advantage is consistent: even its worst run exceeds the baseline's best, and a low standard deviation with closely aligned mean and median rules out lucky outliers (Table~\ref{tab:perf_summary}).

\begin{table}[!htb]
    \caption{Performance summary at $N_e=13$ for each condition under \texttt{Perf-\allowbreak{}Weighted} ($F_1$-based) and naive \texttt{Avg} aggregation (30 runs; all values in \%).}
    \label{tab:perf_summary}
    \centering
    \scriptsize
    \begin{tabularx}{\textwidth}{@{}>{\raggedright\arraybackslash}X rrr rrr@{}}
    \toprule
    & \multicolumn{3}{c}{\textbf{\texttt{Perf-Weighted}}} & \multicolumn{3}{c}{\textbf{\texttt{Avg}}} \\
    \cmidrule(lr){2-4}\cmidrule(lr){5-7}
    \textbf{Method} & \textbf{Mean$\pm$SD} & \textbf{Med$\pm$MAD} & \textbf{Best/Worst} & \textbf{Mean$\pm$SD} & \textbf{Med$\pm$MAD} & \textbf{Best/Worst} \\
    \midrule
    \multicolumn{7}{@{}l}{\textit{\textbf{Independent MoE}}} \\
    Default & \textbf{43.07$\pm$2.45} & \textbf{42.75$\pm$1.75} & 49.00/\textbf{38.00} & 32.93$\pm$3.29 & 32.50$\pm$2.50 & 39.00/27.00 \\
    SBG & 41.23$\pm$3.65 & 41.25$\pm$1.75 & \textbf{50.00}/35.00 & \textbf{35.60$\pm$3.31} & \textbf{36.00$\pm$2.00} & \textbf{42.00}/\textbf{29.00} \\
    SBE & 41.90$\pm$3.26 & 41.25$\pm$1.75 & \textbf{50.00}/35.50 & 29.60$\pm$2.24 & 29.75$\pm$1.25 & 33.50/24.50 \\
    \multicolumn{7}{@{}l}{\textit{\textbf{Shared MoE}}} \\
    Default & 26.03$\pm$2.06 & 25.50$\pm$1.25 & 30.50/22.00 & 22.52$\pm$1.03 & 22.50$\pm$0.50 & 24.00/20.00 \\
    SBG & 23.88$\pm$1.50 & 23.75$\pm$1.00 & 27.00/21.00 & 23.15$\pm$2.57 & 23.25$\pm$1.75 & 28.50/19.00 \\
    SM & 16.75$\pm$2.20 & 17.00$\pm$2.00 & 21.00/12.50 & 13.93$\pm$2.64 & 13.75$\pm$1.75 & 19.00/9.00 \\
    SBG+SM & 19.22$\pm$2.50 & 19.50$\pm$1.75 & 25.50/15.00 & 13.27$\pm$1.84 & 13.50$\pm$1.00 & 16.50/8.50 \\
    \multicolumn{7}{@{}l}{\textit{\textbf{Baseline}}} \\
    Monolithic & 20.95$\pm$2.41 & 20.75$\pm$1.25 & 29.00/17.00 & 20.95$\pm$2.41 & 20.75$\pm$1.25 & 29.00/17.00 \\
    \bottomrule
    \end{tabularx}
\end{table}

\subsection{How Aggregation Strategy Determines Success}
\begin{figure}[!htb]
    \centering
    \includegraphics[width=\textwidth]{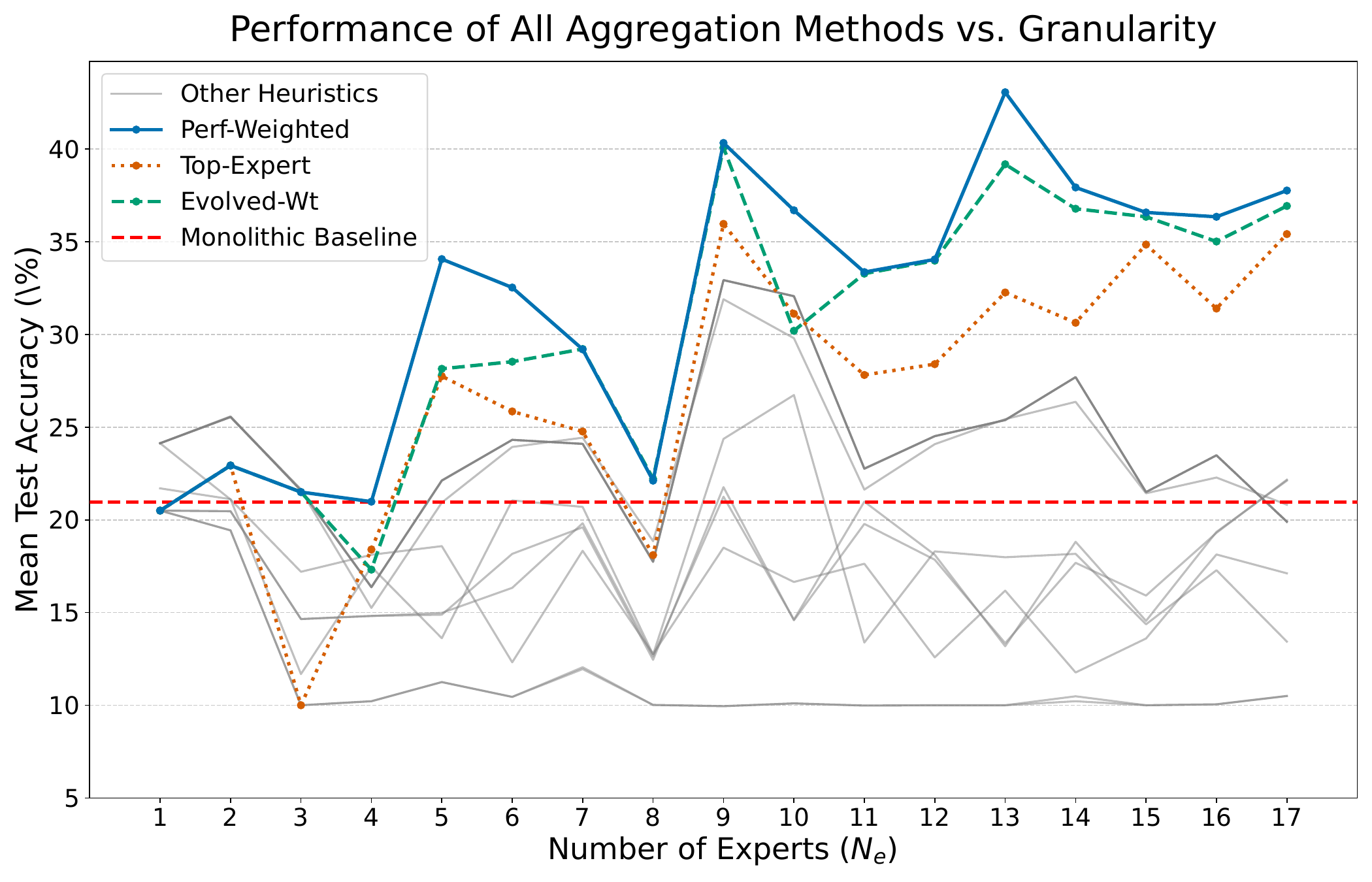}
    \caption{Impact of aggregation strategy on performance across expert counts. Three top $F_1$-weighted strategies (color); the remaining eleven methods (gray).}
    \label{fig:full_aggregation_comparison}
\end{figure}

Expert granularity ($N_e$) matters; aggregation strategy determines how much of the architectural gain reaches the final accuracy. Figure~\ref{fig:full_aggregation_comparison} shows clear stratification across the 14 aggregation methods (one-way ANOVA: $F(13, 406) = 798.06$).

Three $F_1$-weighted strategies outperform the baseline consistently: \texttt{Perf-\allowbreak{}Weighted}, \texttt{Top-Expert}, and \texttt{Evolved-Wt}. These improve with granularity by exploiting expert specialization, peaking at $N_e=10$--13. Simpler heuristics (\texttt{Avg}, \texttt{Sum}, \texttt{Max}) slightly exceed the baseline but behave erratically as expert count grows; naive aggregation cannot resolve conflicting predictions.

The gap is large: at $N_e=13$, \texttt{Perf-\allowbreak{}Weighted} reaches 43.07\% vs.\ 32.93\% for \texttt{Avg} ($t = 13.54$, $d = 3.50$). Yet even with naive \texttt{Avg}, the best \textit{Independent MoE} achieved 35.60\% (Table~\ref{tab:perf_summary}), a 70\% improvement over baseline. Data-driven aggregation maximizes the gain, but the partitioned architecture itself accounts for much of it.

\subsection{Receptive Field Analysis}
\begin{figure}[htbp]
    \centering
    \includegraphics[width=\textwidth]{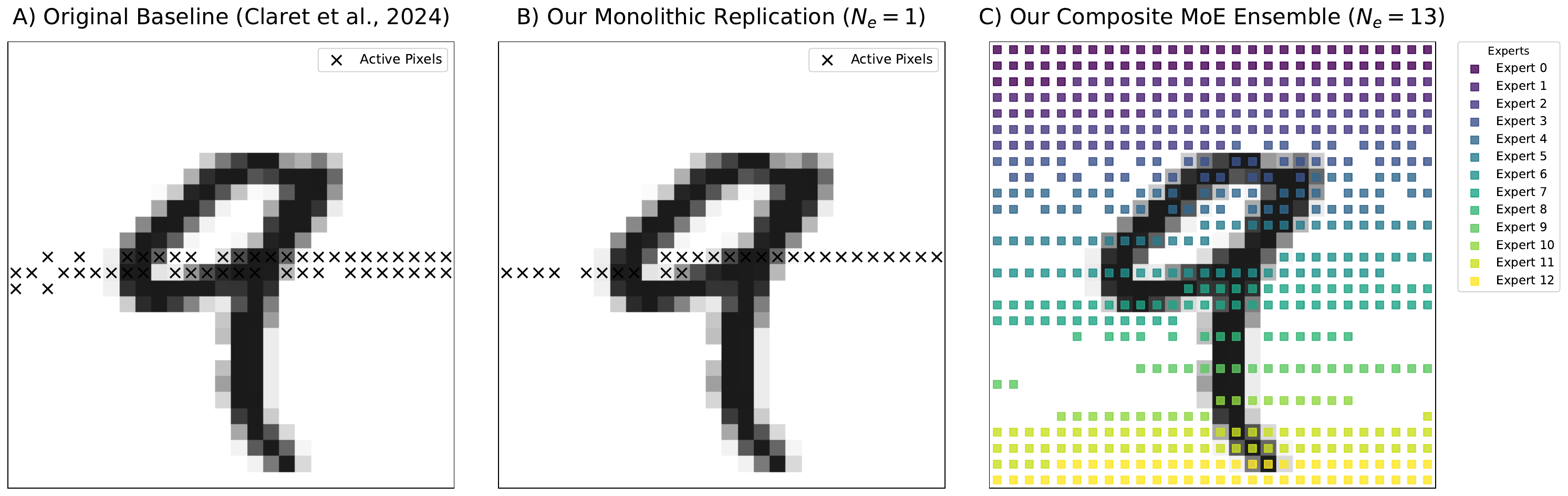}
    \caption{Direct comparison of active receptive fields. \textbf{(A)} The original baseline network from~\cite{claret2024investigating}, with a central focus. \textbf{(B)} Our monolithic ($N_e=1$) replication, reproducing the same centrally biased behavior. \textbf{(C)} The composite field of our 13-expert ensemble (colored squares), with near-complete coverage.}
    \label{fig:receptive_field_comparison_3panel}
\end{figure}

Figure~\ref{fig:receptive_field_comparison_3panel} compares the evolved receptive fields directly. Panel~A: the original monolithic baseline, active pixels confined to a sparse central cluster. Panel~B: our $N_e=1$ replication, faithfully reproducing the same central bias, validating the experimental framework. Panel~C: the composite field of the 13-expert ensemble, with near-complete input coverage.

The original monolithic baseline converged to approximately 48 active pixels. Although each expert sees only its $\approx 60$-pixel segment, the 13 experts together activate over ten times as many pixels as the monolithic baseline.

Coverage growth across expert counts is not strictly monotonic. At $N_e=1$, the model uses only 28 pixels (3.6\% coverage). By $N_e=3$, coverage jumps to 578 pixels (73.7\%). Splitting the 784-pixel input into two halves still leaves each segment well above the capacity ceiling identified earlier in this section, so the qualitative transition from sparse, central convergence to broad coverage sits between two and three experts, not at the very first split. Even minimal partitioning past that crossover breaks the monolithic tendency toward sparse, central solutions. Peak coverage of 78.6\% occurs at $N_e=11$, while our highest-accuracy model ($N_e=13$) uses slightly fewer pixels (71.7\%), suggesting a trade-off between broad coverage and refined feature selection for generalization.

The architectural constraint prevents premature convergence, forcing evolution to discover features across the entire visual field. This mechanism is a major contributor to the \textit{Independent MoE}'s accuracy gains; an alternative CPPN-irregularity hypothesis~\cite{clune2011performance} is discussed in Section~\ref{sec:discussion} and is not refuted by the present data.

\section{Discussion}
\label{sec:discussion}

Decomposing a monolithic neuroevolutionary model into spatially constrained specialists improves performance, and the improvement is further amplified by data-driven aggregation of expert outputs. The \textit{Independent MoE}'s superiority over both the baseline and the \textit{Shared MoE} raises questions about how to evolve solutions for high-dimensional tasks. Our contribution is not the absolute accuracy; a linear classifier exceeds 90\% on MNIST. It is the mechanistic diagnosis: central bias is an architectural limitation of coordinate-based indirect encodings, not a failure of evolutionary search. Partitioning eliminates the bias. This diagnostic approach applies wherever coordinate-based encodings are used on high-dimensional inputs.

\subsection{Architecture and Aggregation: Two Independent Factors}
Two factors drive performance. Architecturally, the \textit{Independent MoE}'s parallel specialization enables focused evolutionary search on each partition, following cooperative coevolution principles, whereas the \textit{Shared MoE}'s sequential processing without partition identity bottlenecks adaptation. The optimal $N_e=13$ balances partition granularity against expert capacity, and Performance-Weighted aggregation resolves the conflicting expert votes that naive schemes cannot as specialist count grows.

Architecture and aggregation contribute independently: even validation-free \texttt{Avg} aggregation clears the baseline by a wide margin (Section~\ref{sec:results}), so data-driven \texttt{Perf-\allowbreak{}Weighted} aggregation amplifies the partitioning effect rather than creating it.

The Default \textit{Independent MoE} (43.07\%) outperforms the SBG variant (41.23\%; $t = 2.29$, $d = 0.59$); the difference with SBE (41.90\%) is not statistically significant ($p = .122$). When each individual is evaluated on a different random batch, fitness-landscape noise helps the population escape local optima. Varied evaluation also implicitly selects for generalization: individuals must perform well across data samples rather than memorizing a fixed batch.

\subsection{Spatial Decomposition vs.\ Ensembling}
The performance gains might stem from spatial partitioning specifically, or simply from combining multiple models. The structural data support spatial partitioning: across all 30 runs of the $N_e{=}1$ baseline replication under the fixed-batch SBE protocol, every active pixel falls within rows 10--16 of the 28$\times$28 image, a central band spanning only 25\% of the height, and the union of all 30 receptive fields covers just 169 of 784 pixels (21.6\%).

Ensemble theory establishes that ensemble benefit requires diversity among members~\cite{krogh1994neural}: ensemble error equals average individual error minus a diversity term. When all members converge to the same central pixels, this diversity term approaches zero, and the ensemble reduces to a single model. The \textit{Independent MoE}'s advantage is therefore not ensemble size but \textit{forced spatial diversity}: each expert must discover features in its assigned partition, producing genuinely complementary specialists.

The \textit{Shared MoE} corroborates this: processing the partitions without partition-identity input, it reaches only 26\% (vs.\ 43\%), showing that spatial diversity alone does not suffice without independent per-partition specialization.

\subsection{On the Nature of Central Bias}
The three-compounding-mechanism account of Section~\ref{sec:background} is a hypothesis we do not directly test. One illustration: CPPN bias inputs can in principle shift symmetric activation responses off-center, but evolution must discover this, and the search space may not reward such shifts when central pixels already provide a locally sufficient solution.

An alternative explanation, drawing on the regularity-performance findings of~\cite{clune2011performance}, is that CPPN-based indirect encodings degrade as target-pattern regularity decreases, a regime complex perceptual tasks may occupy; partitioning would then help by shrinking each expert's substrate, and central bias would be a surface symptom of CPPN-irregularity. Our experiments cannot fully refute this, but they are inconsistent with it being the sole driver: both architectures reduce per-call substrate size, yet only the \textit{Independent MoE} captures the large gain. (The \textit{Shared MoE}'s single re-applied CPPN is a further disadvantage, so this comparison is not clean.) The regularity ceiling nonetheless remains a real limit on how far any partitioning approach can be pushed, since each per-expert CPPN inherits it.

This failure mode, and hence the diagnostic and the partitioning remedy that address it, is specific to coordinate-based indirect encodings and their geometric substrates, not to neuroevolution in general.

\subsection{Limitations and Future Work}
The strongest limitation is that the diagnosis rests on a single dataset. MNIST was chosen as a diagnostic, not a competitive benchmark: its centered digits give a clean, reproducible instance of the failure mode, but that same centering prevents separating the coordinate-system bias from MNIST's information-density bias on MNIST alone. Generalizing the mechanism requires datasets whose informative regions are spatially off-center, which we leave to future work.

Three extensions follow directly from the present analysis.

\textbf{2D / quadrant / block partitioning.} Our 1D linear slicing was deliberately minimal: chosen to isolate ``any partitioning at all'' from ``geometrically informed partitioning.'' 2D rectangular, quadrant, or radial partitionings respect digit geometry. The 106\% gain we report with 1D partitioning should be read as a lower bound on what geometry-preserving partitioning could achieve. A further step is to evolve the partition boundaries jointly with the experts, rather than fixing them a priori, letting the search allocate input area where discriminative features lie. The dual question is per-expert utility: scoring each expert's marginal contribution and pruning redundant or near-silent partitions would test whether all $N_e$ experts earn their place.

\textbf{Plain HyperNEAT baseline.} Plain HyperNEAT uses a fixed substrate with no quadtree, so it has no variance-driven central bias. Comparing it against monolithic ES-HyperNEAT on MNIST is the cleanest discriminator from the CPPN-irregularity hypothesis~\cite{clune2011performance}: if plain HyperNEAT also fails near 21\%, CPPN smoothness is the dominant ceiling; if it does measurably better, central bias is the ES-specific failure mode we claim.

\textbf{Mechanism-isolating ablations.} Cheaper interventions could test which sub-mechanism dominates: non-symmetric-only activation palettes, variance-threshold annealing, explicit center-shifting in CPPN bias inputs, and zeroing default output connections. These would isolate which factor produces the bias rather than only quantifying its cost.

Beyond these three, per-expert hyperparameter optimization could tune each specialist to its segment, and biomedical imagery, where peripheral coverage is clinically informative, is a natural larger-scale target.

\section{Conclusion}
\label{sec:conclusion}

The spatially partitioned \textit{Independent MoE} achieves 43.07\% mean accuracy on MNIST, a 106\% improvement over the 20.95\% monolithic ES-HyperNEAT baseline, and outperforms the \textit{Shared MoE} that processes the same partitions without specialization. Partitioning carries most of the gain: equal-weighted averaging alone yields a 70\% relative improvement, with data-driven aggregation amplifying it further. Receptive-field analysis shows that partitioning expands pixel coverage from 3.6\% to 78.6\%, and forced spatial diversity, rather than ensemble size, drives the improvement. Two tools follow for bio-inspired pattern-recognition systems on high-dimensional inputs: spatial decomposition into evolutionarily discovered specialists to restore receptive-field coverage when monolithic designs collapse, and receptive-field analysis to expose coordinate-bias limitations that accuracy alone would miss.

\noindent Code, configurations, and data to reproduce all experiments are available at \url{https://github.com/RomainClaret/es-hyperneat-optimization-studies}.

\bibliographystyle{splncs04}
\bibliography{references}
\end{document}